\documentclass{llncs}
\usepackage{makeidx}  % allows for indexgeneration
\usepackage{url}

\usepackage[cmex10]{amsmath}
\usepackage{graphicx}
\usepackage{epsfig}
\usepackage{amssymb}

\usepackage{array}

\begin{document}
\frontmatter          % for the preliminaries
\pagestyle{headings}  % switches on printing of running heads
\mainmatter              % start of the contributions

\def\x{{\mathbf x}}
\def\L{{\cal L}}

\title{Spatiotemporal Manifold Prediction Model for Anterior Vertebral Body Growth Modulation Surgery in Idiopathic Scoliosis}
\titlerunning{}  % abbreviated title (for running head)
\author{William Mandel\inst{1} \and Olivier Turcot\inst{2} \and Dejan Knez\inst{3} \and\\Stefan Parent\inst{2} \and Samuel Kadoury\inst{1,2}}
\institute{MedICAL, Polytechnique Montreal, Montreal, QC, Canada \and Sainte-Justine Hospital Research Center, Montreal, QC, Canada \and University of Ljubljana, Faculty of Electrical Engineering, Slovenia \thanks{Supported by the Canada Research Chairs and NSERC Discovery Grants.}}
\authorrunning{Mandel et al.}

%\author{Samuel Kadoury\inst{1,2} \and Hubert Labelle\inst{2} \and Stefan Parent\inst{2}}
%\institute{MEDICAL, Polytechnique Montreal, Montreal, Canada \and Sainte-Justine Hospital Research Center, Montreal, Canada}
%
\authorrunning{me}   % abbreviated author list (for running head)
%
%%%% modified list of authors for the TOC (add the affiliations)

\tocauthor{author1 (Univesity of Author1),
author2 (University of Author2),
author3 (University of Author3),
}

\maketitle              % typeset the title of the contribution

\begin{abstract}

Anterior Vertebral Body Growth Modulation (AVBGM) is a minimally invasive surgical technique that gradually corrects spine deformities while preserving lumbar motion. However the selection of potential surgical patients is currently based on clinical judgment and would be facilitated by the identification of patients responding to AVBGM prior to surgery. We introduce a statistical framework for predicting the surgical outcomes following AVBGM in adolescents with idiopathic scoliosis. A discriminant manifold is first constructed to maximize the separation between responsive and non-responsive groups of patients treated with AVBGM for scoliosis. The model then uses subject-specific correction trajectories based on articulated transformations in order to map spine correction profiles to a group-average piecewise-geodesic path. Spine correction trajectories are described in a piecewise-geodesic fashion to account for varying times at follow-up exams, regressing the curve via a quadratic optimization process. To predict the evolution of correction, a baseline reconstruction is projected onto the manifold, from which a spatiotemporal regression model is built from parallel transport curves inferred from neighboring exemplars. The model was trained on 438 reconstructions and tested on 56 subjects using 3D spine reconstructions from follow-up exams, with the probabilistic framework yielding accurate results with differences of $2.1\pm0.6^{o}$ in main curve angulation, and generating models similar to biomechanical simulations.

\end{abstract}

\section{Introduction}
\label{sec:intro}

Spinal morphology and more particularly 3D morphometric parameters, have demonstrated significant potential in assessing the risk of spinal disease progression. For spinal deformities such as adolescent idiopathic scoliosis (AIS), personalized 3D reconstructions generated from radiographs allows surgeons to assess the severity and decide on efficient treatment options.  A recently introduced minimally invasive surgical technique called Anterior Vertebral Body Growth Modulation (AVBGM) consists of instrumenting the spine with traditional vertebral implants to link a segment of vertebrae together by a flexible polypropylene cable applied to the spine anteriorly. The fusion-less technique applies compressive forces on the convex side of the spinal curve, thereby modulating the distribution of pressure on the vertebral growth plates. In combination with natural bone growth, this allows to retain spine flexibility \cite{skaggs2014classification}. While this new surgical technique showed promising results for skeletally immature patients \cite{crawford2010growth}, difficulties were reported to predict short and long-term post-operative correction \cite{samdani2015anterior}. Biomechanical models were shown to reproduce surgical outcomes, but are not adapted for real-time surgical applications \cite{cobetto20183d}.

A recent study evaluated differences in hand-crafted 3D parameters in progressive AIS groups using images from the patient's first visit  \cite{nault2014three} to predict progression, by manually selecting the best features that can characterize the intrinsic nature of 3D spines. On the other hand, dimensionally-reduced growth trajectories of various anatomical sites have been investigated in neurodevelopment studies for newborns, based on geodesic shape regression to compute the diffeomorphisms based on image time series of a population \cite{singh2013hierarchical}. These regression models were also used to estimate spatio-temporal evolution of the cerebral cortex, by automatically identifying the points of interest and inertia between the first and follow-up images based on non-rigid transformations \cite{fishbaugh2014geodesic}. The concept of parallel transport curves in the tangent space from low-dimensional manifolds proposed by Schiratti et al. \cite{schiratti2015learning} was used to analyze shape morphology \cite{kadoury20173} and adapted for radiotherapy response \cite{chevallier2017learning}, but lacks the capability to predict correction from applied forces following surgery.
% . Regression models were proposed for both cortical and subcortical structures, with 4D varifold-based learning framework with local topography shape morphing being proposed by Rekik et al. \cite{rekik2016predicting}, yet there is no framework adapted for outcomes in spine pathologies

\begin{figure}[tb]
  \begin{center}
        \includegraphics[height=1.38in] {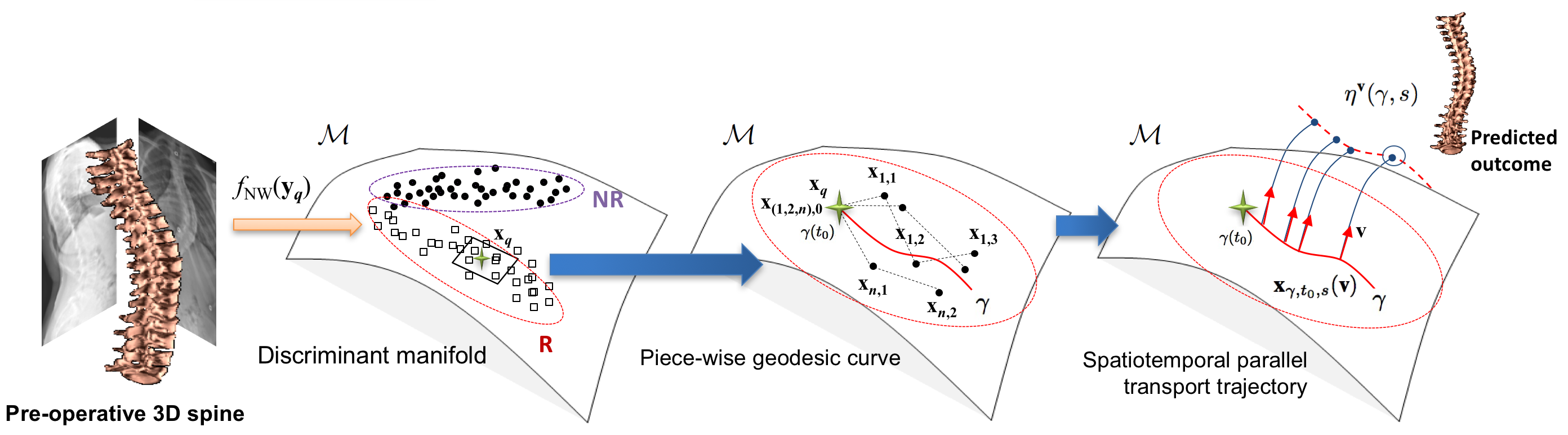}
    \caption{
     Proposed prediction framework for spine surgery outcomes. In the training phase, a dataset of spine models are embedded in a spatio-temporal manifold $\mathcal{M}$, into responsive (R) or non-responsive (NR) groups. During testing, an unseen baseline 3D spine reconstruction $\textbf{y}_q$ is projected on $\mathcal{M}$ using $f_{\text{NW}}$ based on Nadaraya-Watson kernels. The closest samples to the projected point $\textbf{x}$ are selected to regress the spatiotemporal curve $\gamma$ used for predicting the correction due with AVBGM.}
    \label{fig:Flow}
  \end{center}
\end{figure}

This paper presents a prediction model for patient response to AVBGM from pre-operative 3D spine models reconstructed from biplanar X-ray images (Fig. \ref{fig:Flow}). The method first trains a piecewise-geodesic manifold using a collection of pre-operative and longitudinal 3D reconstructions of the spine acquired during follow-up evaluations of patients treated with AVBGM for AIS. A discriminant adjacency matrix is constructed to separate responding and non-responding patients.
% while minimizing the distance in latent variables of samples belonging to the same class. 
During testing, an unseen baseline spine model is projected onto the manifold, where a piecewise-geodesic curve describing spatiotemporal evolution is regressed using discrete approximations, from which the curvature evolution is inferred, yielding a prediction of the intervertebral displacements and shape morphology describing deformation correction.   The main contribution of this paper is the introduction of a piecewise-geodesic transport curve in the tangent space from low-dimensional samples designed for the correction of spinal deformities, where a new time-warping function controlling the rate of correction is obtained from clinical parameters.
%The method was tested on 56 surgical patients with longitudinal 3D spine reconstructions obtained at follow-up.
\section{Method}
\subsection{Discriminant Embedding of Longitudinal Spine Models}
\label{sec.articulatedrep}
A sample spine reconstruction is represented by $\textbf{S}= \{\mathbf{s}_{1},\ldots,\mathbf{s}_{m}\}$, modelling a series of $m=17$ vertebral shapes. For each $\mathbf{s}_{i}$ vertebra, template-based models are obtained where vertex coordinates have one-to-one correspondences between samples. In addition to the mesh-based representation, each model $\mathbf{s}_{i}$ possesses a list of annotated  landmarks to compute local intervertebral rigid transformations, such that $A = [T_{1}, T_{2}, \ldots , T_{m}]$, with $T_i=\{R,t\}$ a rigid inter-vertebral transform. Hence, the overall shape of the spine is described as a vector of sequential registrations assigned to each vertebral level, whereby considering the ensemble of transformations, we obtain a combination of previous transforms:
\begin{equation}\label{eq.Aabs}
\textbf{y}_{\textrm{i}} = [T_{1}, \mathbf{s}_{1}; T_{1} \circ T_{2}, \mathbf{s}_{2},\ldots , T_{1} \circ T_{2} \circ \ldots \circ T_{m},\mathbf{s}_{m}]
\end{equation}
using recursive compositions. The feature array $\textbf{y}_{\textrm{i}}$ dictates the location and rotation of the object constellation, while procuring the morphology of the vertebrae  $\textbf{S}$. The model is deformed by applying displacements to the inter-vertebral parameters. By extending this to the entire absolute vector representing the spine model, this then achieves a global deformation. In this case, registrations are described in the reference coordinate system of the lower vertebra, corresponding to it's principal axes of the cuneiform shape with the origin positioned at the center of mass of the vertebra. The rigid transformations are the combination of a rotation matrix $R$ and a translation vector $t$. We formulate the rigid transformation $T = \{R, t\}$ of a vertebra mesh $\mathbf{s}_{i}$ as $y = Rx+t$ where $x, y, t  \in \Re^{3}$. Composition is given by $T_{1}\circ T_{2}=\{R_{1}R_{2},R_{1}t_{2}+t_{1}\}$.

We propose to embed a collection of non-responsive (NR)  and (2) responsive (R) patients to AVBGM which will offer a maximal separation between the classes, by using a discriminant graph-embedding. Here, $n$ labelled points  $\mathbb{Y}=\{(\textbf{y}_i,l_i,t_i)\}_{i=1}^{n}$ defined in  $\mathbb{R}^{D}$ are embedded in the low-dimensional manifold $\mathcal{M}$, where $l_i$ describes the label (NR or R) and $t_i$ defines the time of follow-up. We assume that for the sampled data, an underlying manifold of the high-dimensional data exists such that  $\mathbb{X}=\{(\textbf{x}_i,l_i,t_i)\}_{i=1}^{n}$ defined in $\mathbb{R}^{d}$. We rely on the assumption that a locally linear mapping $\textbf{M}_i \in \mathbb{R}^{D \times d}$ exists, where local neighbourhoods are defined as tangent planes estimated with ${ \textbf{y}_j - \textbf{y}_i}$ and ${ \textbf{x}_j - \textbf{x}_i}$, describing the paired distances between linked neighbours ${i,j}$. Hence, the relationship can be established as $ \textbf{y}_j - \textbf{y}_i \approx \textbf{M}_i ( \textbf{x}_j - \textbf{x}_i )$.

Because the discriminant manifold structure in $\mathbb{R}^{d}$ requires to maintain the local structure of the underlying data, a undirected similarity graph  $\mathcal{G}=(\textbf{\emph{V}},\textbf{\emph{W}})$ is built, where each node $\textbf{\emph{V}}$ are connected to each other with edges that are weighted with the graph $\textbf{\emph{W}}$. The overall structure of $\mathcal{M}$ is therefore defined with $\textbf{\emph{W}}_w$ for feature vectors belonging to the same class and  $\textbf{\emph{W}}_b$, which separate features from both classes. During the embedding of the discriminant locally linear latent manifold, data samples are divided between $\textbf{\emph{W}}_w$ and $\textbf{\emph{W}}_b$.

\subsection{Piecewise-Geodesic Spatiotemporal Manifold}

Once sample points $\textbf{x}_i$ are in manifold space, the objective is to regress a regular and smooth piecewise-geodesic curve $\gamma : [t_1,t_N]$ that accurately fits the embedded data describing the spatiotemporal correction following AVBGM within a 2 year period. For each sample data $\textbf{x}_i$, the $K$ closest individuals demonstrating similar baseline features are identified from the embedded data, creating neighborhoods $\mathcal{N}(\textbf{x}_q)$ with measurements  at different time points, thus creating a low-dimensional Riemannian manifold where data points $\textbf{x}_{i,j}$, with $i$ denoting a particular individual, $j$ the time-point measurement and $j=0$ the pre-operative model. By assuming the manifold domain is complete and piecewise-geodesic curves are defined for each time trajectories, time-labelled data can be regressed continuously in $\mathbb{R}^{D}$, thereby creating smooth curves in time intervals described by samples in $\mathbb{R}^{d}$.

However, due to the fact the representation of the continuous curve is a variational problem of infinite dimensional space, the implementation follows a discretization process which is derived from the procedure in \cite{boumal2011discrete}, such that:
\begin{align}\label{eq.discreteminimization}
E(\gamma) = \dfrac{1}{K_d} \sum_{i=1}^{K_d}  &\sum_{j=0}^{t_N} w_i \| \gamma(t_{i,j}) - (\textbf{x}_{i,j} - (\textbf{x}_{i,0} - \textbf{x}_q)) \|^2 \nonumber \\
&+ \dfrac{\lambda}{2}  \sum_{i=1}^{K_d} \alpha_i \|v_i\|^2 + \dfrac{\mu}{2}  \sum_{i=1}^{K_d} \beta_i \| a_i \|^2.
\end{align}

This minimization process simplifies the problem to a quadratic optimization, solved with LU decomposition. The piecewise nature is represented by the term $K_d \in \mathcal{N}(\textbf{x}_q)$, defined as samples along $\gamma$. The 1$^{st}$ component of Eq.(\ref{eq.discreteminimization}) is a penalty term to minimize the geodesic distance between samples  $\textbf{x}_{i,j}$  and the regressed curve, where $w_i$ are weight variables based on sample distances. This helps regress a curve that will lie close to $\textbf{x}_{i,j}$, shifted by  $\textbf{x}_q$ in order to have the initial reconstructions co-registered. The 2$^{nd}$ term represents the velocity of the curve (defined by $v_i$, approximating $\dot{\gamma}(t_i)$), minimizing the $L_2$ distance of the 1$^{st}$ derivative of $\gamma$. By minimizing the value of the curve's first derivatives, this prohibits any discontinuities or rapid transitions of the curve's direction, and is modulated by $\alpha_i$. Finally, an acceleration penalty term (defined by  $a_i$) focuses on the 2$^{nd}$ derivative of $\gamma$ with respect to $t_i$ by minimizing the $L_2$ norm. The acceleration is modulated  by $\beta_i$. Estimates for $v_i$ and $a_i$ (weighted by $\{ \lambda,\mu \}$, respectively), are generated using geometric finite differences. These estimates dictates the forward and backward step-size on the regressed curve, leading to directional vectors in $\mathcal{M}$ as shown in \cite{boumal2011discrete}. In order to minimize $E(\gamma)$, a non-linear conjugate gradient  technique defined in the low-dimensional space $\mathbb{R}^{d}$ is used, thus avoiding convergence and speed issues. The regressed curve $\gamma$ is therefore defined for all time points, originating at $t_0$. The curve creates a group average of spatiotemporal transformations based on individual correction trajectories.

\subsection{Prediction of Spine Correction}

Finally, to predict the evolution of spine correction from an unseen pre-operative spine model, we use the geodesic curve $\gamma : \mathbb{R}^{D}  \rightarrow \mathcal{M}$ modelling the spatiotemporal changes of the spine, where each point $\textbf{x} \in \mathcal{M} $ is associated to a speed vector $\textbf{v}$ defined with a tangent plane on the manifold such that $\textbf{v} \in \text{T}_{\textbf{x}} \mathcal{M}$. 

Based on Riemannian theory, an exponential mapping function at $\textbf{x}$ with velocity $\textbf{v}$ can be defined from the geodesics such that $ \text{e}_{\textbf{x}}^{\mathcal{M} }(\textbf{v})$. Using this concept, parallel transport curves defined in $\text{T}_{\textbf{x}}$ can help define a series of time-index vectors along $\gamma$ as proposed by \cite{schiratti2015learning}. The collection of parallel transport curves allows to generate an average trajectory in ambient space $\mathbb{R}^{D}$, describing the spine changes due to the corrective forces of tethering. The general goal is to begin the process at the pre-operative sample, and navigate the piecewise-geodesic curve describing correction evolution in time, where one can extract the appearance at any point (time) in $ \mathbb{R}^{D}$ using the exponential mapping. For implementation purposes, the parallel transport curve are constrained within a smooth tubular boundary perpendicular to the curve (from an ICA) to generate the spatiotemporal evolution in the coordinate system of the pre-operative model.
  
Hence, given the manifold at time $t_0$ with $\textbf{v}$ defined in the tangent plane and the regressed piecewise-geodesic curve $\gamma$, the parallel curve is obtained as:
\begin{equation}\label{eq.parallelcurve}
\eta^{\textbf{v}} (\gamma,s)= \text{e}_{ \gamma(s)}^{\mathcal{M}} (\textbf{x}_{\gamma,t_0, s}(\textbf{v})), \,\,\, s \in \mathbb{R}^{d}.
\end{equation}
Therefore by repeating this mapping for manifold points seen as samples of individual progression trajectories along $\gamma(s)$, an evolution model can be generated. Whenever a new sample is embedded, new samples points along $\gamma(s)$, denoted as $\eta^{\textbf{v}} (\gamma,\cdot)$ can be generated parallel to the regressed piecewise curve in $\mathcal{M}$, capturing the spatiotemporal changes in correction.

A time warp function allowing $s$ to vary along the geodesic curve is described as $\phi_i(t)=  \theta_i (t-t_0-\tau_i) +t_0$. Here, we propose to incorporate a personalized acceleration factor based on the spine maturity and flexibility derived from the spine bending radiographs and Risser grade. A coefficient $\theta_i = C_i \times R_i$ describing the change in Cobb angle $C_i$ between poses, and modulated by the Risser grade $R_i$. This coefficient regulates the rate of correction based on the $K$ neighbouring samples. Finally, to take under account the relative differences between the group-wise samples and the query model once mapped onto the regressed curve, a time-shift parameter  $\tau_i$ is incorporated in the warp function.

For spine correction evolution,  displacement vectors $\textbf{v}_i$ are obtained by a PCA of the hyperplane crossing $\text{T}_{\textbf{x}_i} \mathcal{M}$  in manifold $\mathcal{M}$ \cite{schiratti2015learning}. Hence, for any query sample $\textbf{x}_q$ which represents the mapped pre-operative 3D reconstruction (prior to surgery), the predicted model at time $t_k$ can be regressed from the piecewise-geodesic curve generated from embedded samples $\textbf{x}$ in $\mathcal{N}(\textbf{x}_q)$ such that:
\begin{equation}\label{eq.spatiotemporalmodel}
\textbf{y}_{q,t_k} = \eta^{\textbf{v}_q} (\gamma,\phi_i(t_{k})) + \epsilon_{q,t_k}
 \end{equation}
which yields a predicted post-operative model $\textbf{y}_{q,t_k}$ in high-dimensional space $\mathbb{R}^{D}$, and $\epsilon_{q,t_k}$ a zero-mean Gaussian distribution. The generated model offers a complete constellation of inter-connected vertebral models composing  the spine shape $\textbf{S}$, at first-erect (FE), 1 or 2-year visits, including landmarks on vertebral endplates and pedicle extremities, which can be used to capture the local shape morphology with the correction process.

\section{Experiments}

The discriminant manifold was trained from a database of $438$ 3D spine reconstructions generated from biplanar images \cite{Humbert09}, originating from $131$ patients demonstrating several types of deformities with immediate follow-up (FE), 1 year and 2 year visits. Patients were recruited from a single center prospective study, with the inclusion criteria being evaluated by an orthopaedic surgeon and a main curvature angle between 30$^{\circ}$ and 60$^{\circ}$. Patients were divided in two groups, with the first group composed of 94 responsive patients showing a reduction in Cobb angle over or equal to 10$^{\circ}$ between the FE and follow-up visit. The second group was composed of 37 non-responsive (NP) patients with a reduction of less than 10$^{\circ}$. Each vertebra model of the spine were annotated with 4 pedicle tips and 2 center points placed on the vertebral endplates, and validated by an experienced radiologist. These expert-selected landmarks were used to establish the local coordinate system for each vertebra, describing the orientation and location (known pose), and used as control points to warp triangulated shape models generated from CT images of a cadaveric spine.

\begin{figure}[t!]
  \begin{minipage}[b]{0.46\linewidth}
  \centering
  \includegraphics[height=1.95in] {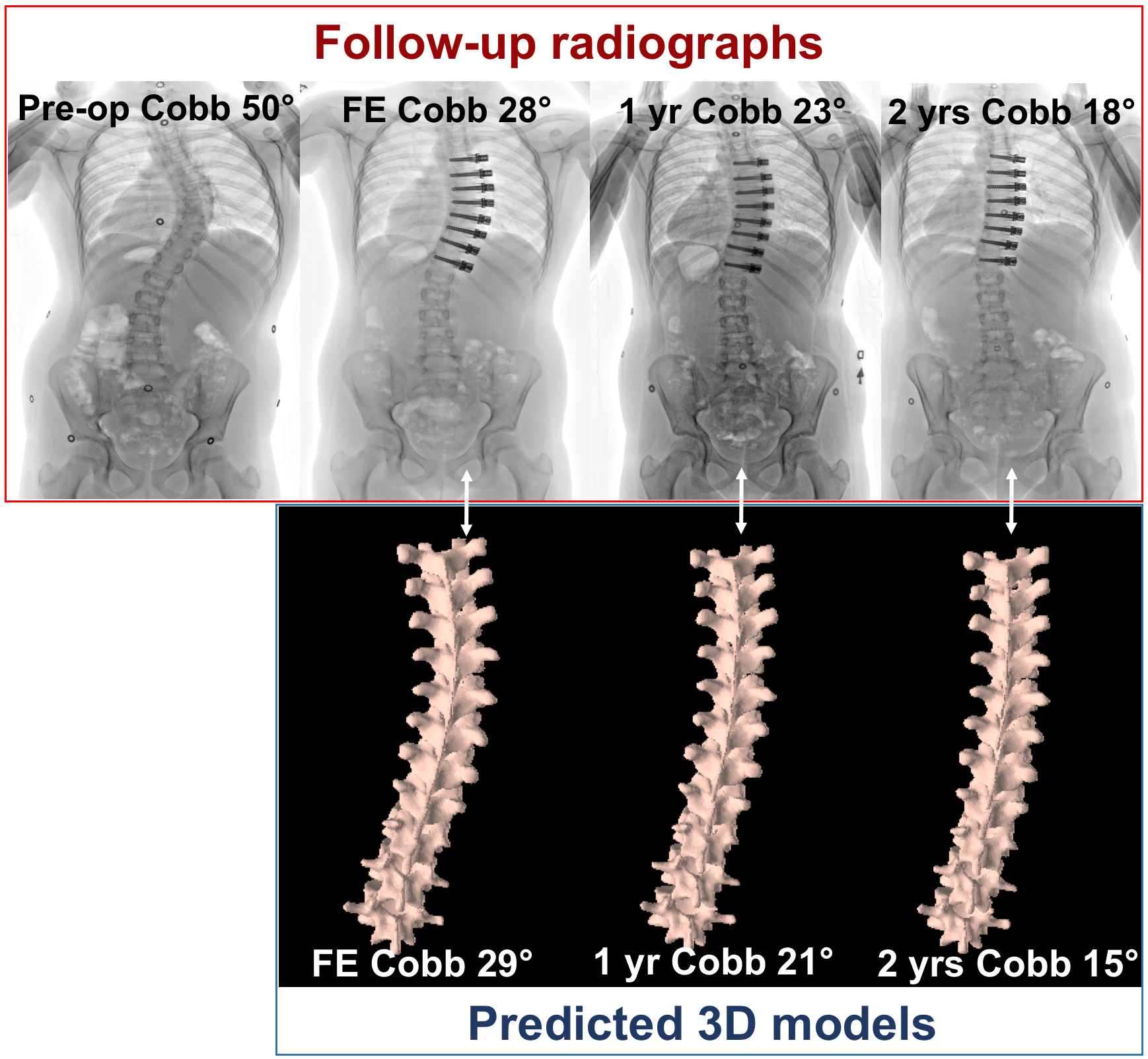}
  \vspace{0.2cm}
  \centerline{  (a)}
\end{minipage}
\begin{minipage}[b]{0.49\linewidth}
  \centering
  \includegraphics[height=1.8in, width=2.7in] {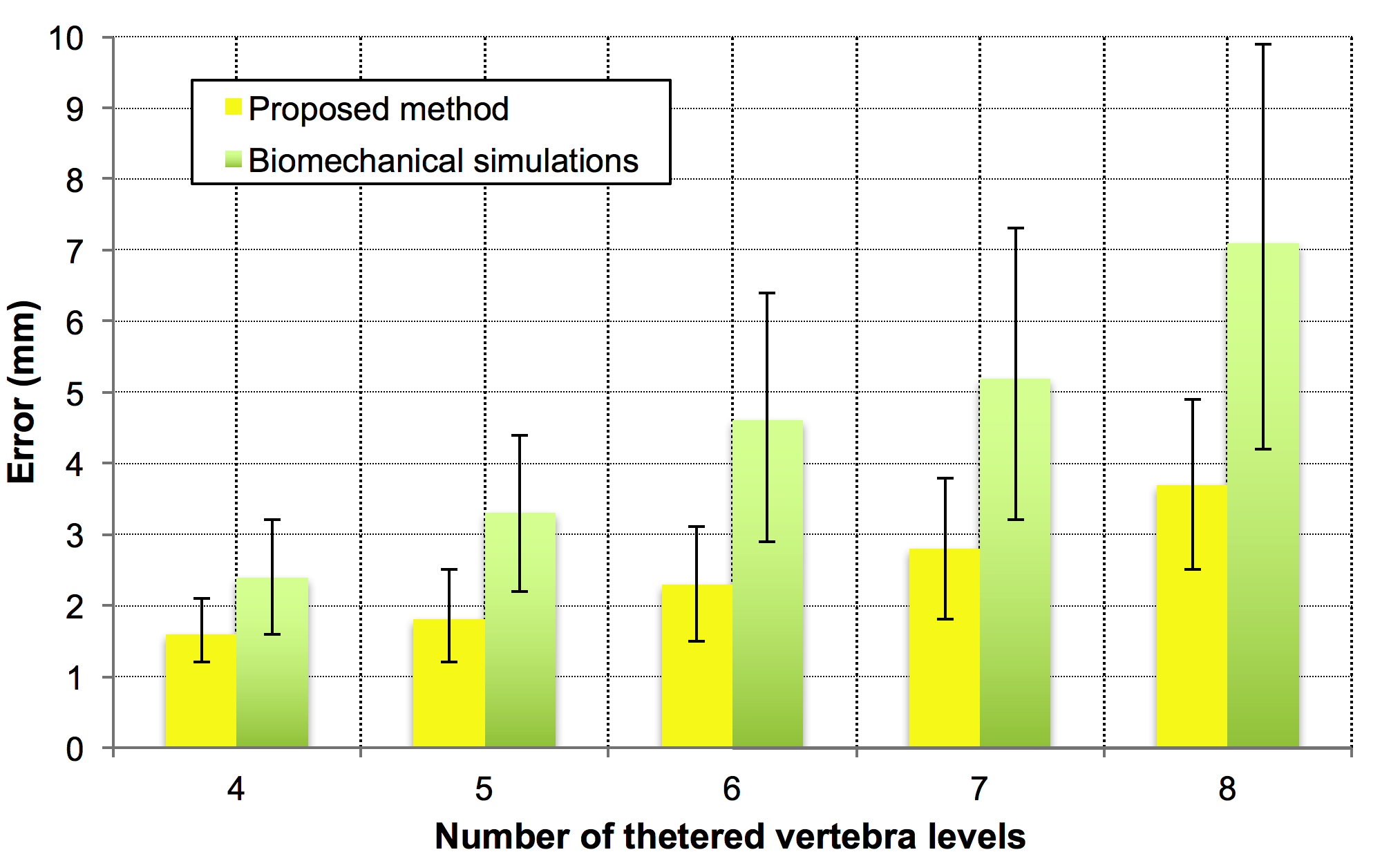}
  \vspace{0.2cm}
  \centerline{(b)}
\end{minipage}

\caption{(a) Comparison in actual and predicted Cobb angles in a 11 y.o. patient at the first-erect visit, at 1-yr and at 2-yrs postop. Top row depicts the actual X-rays, while the bottom row presents the predicted 3D spine geometry. (b) Errors with 5 different tethering levels, comparing results with biomechanical simulations at 2 yrs.}
\label{fig.samples}
\end{figure}

We evaluated the geometrical accuracy of the predictive manifold for 56 unseen surgical patients with AVBGM (mean age $12 \pm 3$, average main Cobb angle on the frontal plane at the first visit was $47^{\circ} \pm 10^{\circ}$), with predictions at $t=0$ (FE), $t=12$ and $t=24$ months. For the predicted models, we evaluated the 3D root-mean-square difference of the vertebral landmarks generated, the Dice coefficients of the vertebral shapes and  in the main Cobb angle. The results are shown in Table \ref{table_example}. Results were confronted to other techniques such as biomechanical simulations performed on each subject using finite element modelling with ex-vivo parameters \cite{cobetto20183d}, a locally linear latent variable model \cite{park2015bayesian} and a deep auto-encoder network \cite{thong2016three}. Fig. \ref{fig.samples} shows a sample prediction result for an 11 y.o. patients at FE, 12 and 24-months for a patient with right-thoracic deformity, which are more common in the scoliotic population. Results from the predicted geometrical models show the regressed spatio-temporal geodesic curve yields anatomically coherent structures, with accurate local vertebral morphology.

To evaluate robustness with respect to varying instrumented levels, we measured the accuracy of the predicted models for tethering between 4 and 8 vertebrae at 2 yrs, ranging from thoracic to lumbar regions. Fig. \ref{fig.samples}(b) shows the improvement of the spatiotemporal geodesic curve in comparison to traditional biomechanical models, particularly when the number of levels are higher.

\section{Conclusion}
\label{sec:discussion}
In this paper, we proposed an accurate predictive model of spine morphology and Cobb angle correction obtained at the first-erect, 1-year and 2-year visits, following anterior vertebral body growth modulation. The piecewise-geodesic curve capturing spatio-temporal changes could be used for patient selection of AVBGM as a decision-sharing tool prior to surgery. Our approach is based on smooth and regular trajectories embedded in a discriminant manifold, which enable an efficient navigation on a low-dimensional domain trained from operative cases, yielding results similar to actual surgical outcomes. Future work will include a multi-center evaluation before it can be used in clinical practice.

\begin{table*}[!t]

\renewcommand{\arraystretch}{1}
\caption{3D RMS errors (mm), Dice (\%) and Cobb angles ($^{o}$) for the proposed method, and compared with biomechanical simulations, locally linear latent variable models (LL-LVM) and deep auto-encoders (AE). Predictions are evaluated at FE, 1 and 2-yrs.}
\label{table_example}
\centering
\scalebox{0.95}{
\begin{tabular}{|l||c|c|c|c|c|c|c|c|c|}
\hline

 & \multicolumn{3}{c|}{FE visit} & \multicolumn{3}{c|}{1-year visit} & \multicolumn{3}{c|}{2-year visit}  \\
    
\cline{2-10}
 & 3D RMS  & Dice & Cobb & 3D RMS   & Dice & Cobb   &  3D RMS  & Dice & Cobb   \\
\hline
Biomec. sim &  	3.3$\pm$1.1 & 85$\pm$3.4 &   2.8$\pm$0.8 & 		3.6$\pm$1.2 & 84$\pm$3.6 & 3.2$\pm$0.9 & 		4.1$\pm$2.3 &  82$\pm$3.9 & 3.6$\pm$1.0  \\
LL-LVM  \cite{park2015bayesian} & 		3.6$\pm$1.4 & 83$\pm$4.0 &   3.8$\pm$1.5 & 		4.7$\pm$3.3 & 79$\pm$4.4 & 5.5$\pm$2.6 & 		6.6$\pm$4.4 &  71$\pm$5.9 & 7.0$\pm$3.9  \\
Deep AE \cite{thong2016three} &  	4.1$\pm$1.5 & 80$\pm$4.4 &   5.1$\pm$2.7 & 		5.0$\pm$1.9 & 77$\pm$4.9 & 5.8$\pm$3.0 & 		6.3$\pm$4.6 &  72$\pm$5.7 & 6.6$\pm$4.2 \\
\hline
\textbf{Proposed} &  2.4$\pm$0.8 & 92$\pm$2.7 &   1.8$\pm$0.5 & 2.9$\pm$0.9 & 90$\pm$2.8 & 2.0$\pm$0.7 & 3.2$\pm$1.3 &  87$\pm$3.1 & 2.1$\pm$0.6  \\
\hline
\end{tabular}
}
\end{table*}

\bibliographystyle{splncs}
\bibliography{sample}
\end{document}